\documentclass[preprint,5p,times,twocolumn]{elsarticle}
\usepackage[utf8]{inputenc}
\usepackage{amsmath}
\usepackage{amssymb}
\usepackage{graphicx}
\usepackage{subfigure}
\usepackage{natbib}
\biboptions{sort&compress}
\usepackage{amsthm,amsmath,amssymb}
\usepackage{mathrsfs}
\usepackage{booktabs} 

\usepackage{multirow} 
\usepackage{longtable} 
\usepackage{array} 
\usepackage{makecell}
\usepackage{threeparttable} 
\usepackage{booktabs}

\journal{Knowledge-Based Systems}

\begin{document}

\begin{frontmatter}

\title{Novel deep learning methods for 3D flow field segmentation and classification}

\author[1]{Xiaorui Bai}
\ead{xrbai@std.uestc.edu.cn}

\author[1]{Wenyong Wang}
\ead{wangwy@uestc.edu.cn}

\author[1]{Jun Zhang}
\ead{zhangjun@uestc.edu.cn}
 
\author[2]{Yueqing Wang}
\ead{yqwang2013@163.com}
 
\author[1]{Yu Xiang\corref{cor1}} 
\ead{jcxiang@uestc.edu.cn}
 
\address[1]{School of Computer Science and Engineering, University of Electronic Science and Technology of China, Chengdu, Sichuan, China}
\address[2]{Computational Aerodynamics Institute, China Aerodynamics Research and Development Center, Mianyang, Sichuan, China}
 
\cortext[cor1]{Corresponding author at: School of Computer Science and Engineering, University of Electronic Science and Technology of China, Chengdu, China.}

\date{January 2023}

\begin{abstract}
Flow field segmentation and classification help researchers to understand vortex structure and thus turbulent flow. Existing deep learning methods mainly based on global information and focused on 2D circumstance. Based on flow field theory, we propose novel flow field segmentation and classification deep learning methods in three-dimensional space. We construct segmentation criterion based on local velocity information and classification criterion based on the relationship between local vorticity and vortex wake, to identify vortex structure in 3D flow field, and further classify the type of vortex wakes accurately and rapidly. Simulation experiment results showed that, compared with existing methods, our segmentation method can identify the vortex area more accurately, while the time consumption is reduced more than 50\%; our classification method can reduce the time consumption by more than 90\% while maintaining the same classification accuracy level.

\end{abstract}

\begin{keyword}
Three-dimensional space \sep Deep learning \sep Flow field segmentation\sep Flow field classification
\end{keyword}

\end{frontmatter}

\section{Introduction}

Vortex is a natural phenomenon which plays an important role in aerodynamics, such as turbulent transition, quantum turbulence and energy cascade, etc
\cite{shi2020pod,panico2023onset,carbone2020vortex}. Vortex is closely related to turbulent flow, a common state of flow and an important area of flow field research. Vortex structures directly affect the generation and evolution of turbulence, and the interaction between vortex structure and turbulence is highly consistent in the wake \cite{pradeep2010vortex,yu2021large}. Therefore, flow field segmentation and classification to identify and classify vortex structures is essential for turbulence research.

Flow field segmentation, also known as vortex identification, is to identify and extract vortex structures from flow field to study the behaviour of vortices. It is widely applied in aerodynamics, such as accurate prediction of rotor wake and the construction of DNS (direct numerical simulation) boundary layer database, etc \cite{kenwright1997vortex,pierce2013application}.

In flow field segmentation area, numerical calculation-based methods and deep learning-based methods are the two most common methods \cite{jiang2005detection}.

Methods based on numerical calculation are further divided into local detection methods and global detection methods. Local detection methods such as $Q$-criterion \cite{hunt1987vorticity}, $\omega$-criterion \cite{liu2016new}, $\lambda_2$-criterion \cite{jeong1995identification}, $\Delta$-criterion \cite{chong1990general} and $\Gamma_2$-criterion \cite{graftieaux2001combining}, detected vortex area using local patches and computing new variables. However, local methods rely heavily on the selection of thresholds to produce valid results, and there are still many false positives and false negatives in the test results. Global methods can obtain high-precision results through global numerical calculations. Sadarjoen extracted vortices using the winding angle and the geometry of streamlines \cite{sadarjoen1998selective}; Serra and Haller created an automated method to efficiently compute closed instantaneous curves using geodesic structures \cite{serra2017efficient}; a threshold-independent method was proposed by Haller to identify vortex based on average vorticity and Instantaneous Vorticity Deviation (IVD) value \cite{haller2016defining}. The above global detection methods often have an expensive computational time cost due to the complexity of global computation.

Recently, deep learning methods are widely used in this field \cite{tian2019novel,lin2021multi,chen2022knowledge,hu2022aerodynamic,hu2022flow}. It is proved to be an compromise way to finish the segmentation task \cite{epps2017review}, which combines the advantages of local and global methods. Franz and Lguensat tried to use CNN-based binary classification method to identify ocean eddies \cite{franz2018ocean,lguensat2018eddynet}. Rajendran applied Recurrent Neural Network (RNN) to three-dimensional flow field to locate the eddy current \cite{rajendran2018vortex}. Wang proposed a fast eddy detection method based on fully convolutional segmentation network \cite{wang2021rapid}. Deng designed a Vortex-U-Net network based on the U-Net model to realize global 3D eddy current detection, which used multiple convolution downsampling for feature extraction and deconvolution upsampling to restore 3D labels \cite{deng2022vortex}. The above-mentioned methods guarantee a certain accuracy rate, but still can be improved in terms of time consumption.

Flow field classification is to classify vortex shedding wake modes in fluid, such as "2S", "2P" and "P+S" mode \cite{williamson1988vortex,mawat2018simulation}. It can help understand vortex shedding patterns, which does benefit to vortex shedding suppression and wake control \cite{rashidi2016vortex}.  Besides, flow field classification helps to reduce the unbalanced forces for applications suffering from vortex-induced vibrations \cite{cann2021mode}.

In flow field classification area, classification methods can also be simply summarized into two types: classification methods based on local measurements and global information.

Wang and Hemati used the ideal flow method to generate "2S" and "2P" wakes with point vortices, and then used the k-nearest neighbor (KNN) algorithm to classify the wake types  \cite{wang2019detecting}. Colvert simulated flow fields with different Strouhal numbers \cite{schnipper2009vortex}, and used the periodicity at fixed points in the flow field as the basis for classification to train the network for flow field classification \cite{colvert2018classifying}. Alsalman further investigated how the accuracy of vortex classification is affected by sensor type, and trained a neural network to classify wake types from "one shot" measurement data on a spatially distributed sensor array \cite{alsalman2018training}. Li considered local measurements based on flow variables, and used the flow direction velocity component, transverse velocity component, vorticity and the combination of the three flow variables to train different neural networks to classify the wake structure \cite{li2020classifying}. Ribeiro explored the kinematics of 46 oscillating wings, combining a convolutional neural network (CNN) with a long short-term memory (LSTM) unit \cite{ribeiro2022machine}. At the same time, they used the convolutional autoencoder combined with the k-means++ clustering method to verify the physical wake differences between airfoil kinematics, and obtained four different wake patterns. The above methods are mainly based on different oscillation frequencies or global information to classify vortex wakes. However, classification research mainly focuses on 2D cases, and there is a lack of classification for 3D flow fields. Furthermore, processing global information is quite time consuming.

Novel deep learning methods for 3D flow field segmentation and classification based on flow field theory are proposed in this paper. The main contributions of our work are as follows:

(1) The relationship among local vorticity, Reynolds number and vortex wake is explored to construct 3D flow field classification criterion. Then a novel 3D flow field classification deep learning method with high accuracy and low time consumption based on this criterion is proposed. 

(2) 3D flow field segmentation criterion identifying vortex through local velocity information is also explored in this paper. A novel flow field segmentation deep learning method, to recognize vortices accurately and rapidly under 3D circumstance, is then proposed.

This paper is organized as follows. Section 2 presents some definitions of flow field and forms the flow field segmentation and classification criteria. Section 3 presents the framework of our flow field segmentation and classification methods. In Section 4, our methods are tested on several classical cases and the corresponding experimental results are analyzed and discussed. Finally, Section 5 concludes our work.

\section{Methods}

\subsection{Problem description and Basic concepts}
\label{problem description}

For segmentation and classification methods we mentioned above, efficiency still needs to be further improved while ensuring high accuracy. This problem mainly comes from two aspects: criterion and network.

\emph{1) Segmentation criterion and network}: Segmentation criterion determines the selection of data and design of network structure. Obtaining the mapping between velocity data and vortex region is the core issue in constructing segmentation criterion. For segmentation network, increasing network depth for higher accuracy will result in reduced efficiency. To reduce complexity, simple network structure is considered according to segmentation criterion.

\emph{2) Classification criterion and network}: For classification task, an intuitive criterion is to classify flow field based on global information. However, the processing of global information leads to excessive computation, thus reducing classification efficiency. Moreover, existing classification criteria mainly focus on 2D cases. Therefore new 3D classification criterion needs to be constructed. One crucial problem is to obtain the mapping between vorticity data and vortex wake type. And for classification network, global vorticity should be avoided as input, and thus new network also should be considered through classification criterion.

In order to solve the above problems, we start from the following basic concepts.

Given a flow field attribute set 
$\mathcal{A} = \left\{ \textbf{p},~t,~\textbf{v} \right\}$ in 3D flow field $\mathcal{F}$, each discrete sampling point has its own physical quantities including position $\textbf{p} = \left\{ x,y,z \right\}$, time $t$ and velocity $\textbf{v} = \left\{ u,v,w \right\}$. With Euler's method, the flow velocity at each position $\textbf{p}$ within $\mathcal{F}$ could be defined as a function of space position $\textbf{p}$ and time $t$. Then, the projections of flow velocity in flow field $\mathcal{F}$ are expressed as follows:
$$
\left\{ \begin{matrix}
{u = u\left( {\textbf{p},~t} \right) = u\left( {x,y,z,t} \right)} \\
{v = v\left( {\textbf{p},~t} \right) = v\left( {x,y,z,t} \right)} \\
{w = w\left( {\textbf{p},~t} \right) = w\left( {x,y,z,t} \right)} \\
\end{matrix} \right.
$$

Further, for flow field $\mathcal{F}$, 3D vorticity is define as the curl of velocity $\textbf{v}$ and denoted by symbol $\omega_{3D}$ \cite{wu2007vorticity}. In the Cartesian coordinate system, $\omega_{3D}$ is calculated as follows:
\begin{equation}
\begin{aligned}
\omega_{3D} &= \nabla \times \textbf{v} = \omega_{x}i + \omega_{y}j + \omega_{z}k = \left| \begin{matrix}
i & j & k \\
\frac{\partial}{\partial x} & \frac{\partial}{\partial y} & \frac{\partial}{\partial z} \\
u & v & w \\
\end{matrix} \right|\\ 
& = \left( {\frac{\partial w}{\partial y} - \frac{\partial v}{\partial z}} \right)i + \left( {\frac{\partial u}{\partial z} - \frac{\partial w}{\partial x}} \right)j + \left( {\frac{\partial v}{\partial x} - \frac{\partial u}{\partial y}} \right)k \\
\end{aligned}
\label{vorticity}
\end{equation}
where $\nabla$ is defined as gradient operator.

\subsection{Segmentation criterion}
\label{design_ffs}

Vorticity has an inseparable relationship with vortex region, but it cannot be directly applied to identify vortices in general \cite{liu2019third}. Therefore, it is not feasible to use vorticity to segment the vortex region in our method.

 Existing flow field segmentation criteria \cite{hunt1987vorticity,liu2016new,chong1990general,jeong1995identification} are based on the analysis of the velocity gradient tensor $\nabla v$ expressed in (\ref{Velocity gradient tensor}).

\begin{equation}
\begin{aligned}
\nabla v = \begin{bmatrix}
\frac{\partial u}{\partial x} & \frac{\partial u}{\partial y} & \frac{\partial u}{\partial z} \\
\frac{\partial v}{\partial x} & \frac{\partial v}{\partial y} & \frac{\partial v}{\partial z} \\
\frac{\partial w}{\partial x} & \frac{\partial w}{\partial y} & \frac{\partial w}{\partial z} \\
\end{bmatrix}
\end{aligned}
\label{Velocity gradient tensor}
\end{equation}

Based on the idea of dividing vorticity into vortical part and
non-vortical part \cite{liu2016new}, velocity gradient tensor $\nabla v$ can be rewritten in (\ref{Velocity gradient tensor division}).

\begin{equation}
\nabla v = \frac{1}{2}\left( {\nabla v + \nabla v^{T}} \right) + \frac{1}{2}\left( {\nabla v - \nabla v^{T}} \right) = A + B
\label{Velocity gradient tensor division}
\end{equation}
where $A$ is the symmetric part and $B$ is the anti-symmetric part, respectively expressed as follows:
$$
A = \frac{1}{2}\left( {\nabla v + \nabla v^{T}} \right) = \begin{bmatrix}
\frac{\partial u}{\partial x} & {\frac{1}{2}\left( {\frac{\partial u}{\partial y} + \frac{\partial v}{\partial x}} \right)} & {\frac{1}{2}\left( {\frac{\partial u}{\partial z} + \frac{\partial w}{\partial x}} \right)} \\
{\frac{1}{2}\left( {\frac{\partial v}{\partial x} + \frac{\partial u}{\partial y}} \right)} & \frac{\partial v}{\partial y} & {\frac{1}{2}\left( {\frac{\partial v}{\partial z} + \frac{\partial w}{\partial y}} \right)} \\
{\frac{1}{2}\left( {\frac{\partial w}{\partial x} + \frac{\partial u}{\partial z}} \right)} & {\frac{1}{2}\left( {\frac{\partial w}{\partial y} + \frac{\partial v}{\partial z}} \right)} & \frac{\partial w}{\partial z} \\
\end{bmatrix}
\label{Symmetric part}
$$

$$
B = \frac{1}{2}\left( {\nabla v - \nabla v^{T}} \right) = \begin{bmatrix}
0 & {\frac{1}{2}\left( {\frac{\partial u}{\partial y} - \frac{\partial v}{\partial x}} \right)} & {\frac{1}{2}\left( {\frac{\partial u}{\partial z} - \frac{\partial w}{\partial x}} \right)} \\
{\frac{1}{2}\left( {\frac{\partial v}{\partial x} - \frac{\partial u}{\partial y}} \right)} & 0 & {\frac{1}{2}\left( {\frac{\partial v}{\partial z} - \frac{\partial w}{\partial y}} \right)} \\
{\frac{1}{2}\left( {\frac{\partial w}{\partial x} - \frac{\partial u}{\partial z}} \right)} & {\frac{1}{2}\left( {\frac{\partial w}{\partial y} - \frac{\partial v}{\partial z}} \right)} & 0 \\
\end{bmatrix}
\label{Antisymmetric part}
$$

From (\ref{Velocity gradient tensor division}), the first symmetric part is related to deformation while the second anti-symmetric part is related to vorticity \cite{liu2016new}. The squares of Frobenius norm of $A$ and $B$ are then introduced as given in (\ref{Frobenius norm}).

\begin{equation}
\begin{aligned}
a = trace\left( {A^{T}A} \right) = {\sum_{i = 1}^{3}{\sum_{j = 1}^{3}\left( A_{ij} \right)^{2}}}\\
b = trace\left( {B^{T}B} \right) = {\sum_{i = 1}^{3}{\sum_{j = 1}^{3}\left( B_{ij} \right)^{2}}}
\label{Frobenius norm}
\end{aligned}
\end{equation}

Define vector $\textbf{s}$ in (\ref{segmentation input}) from velocity gradient tensor $\nabla v$ in (\ref{Velocity gradient tensor}).

\begin{equation}
\textbf{s} = \left\lbrack \frac{\partial u}{\partial x},\frac{\partial u}{\partial y},\frac{\partial u}{\partial z},\frac{\partial v}{\partial x},\frac{\partial v}{\partial y},\frac{\partial v}{\partial z},\frac{\partial w}{\partial x},\frac{\partial w}{\partial y},\frac{\partial w}{\partial z} \right\rbrack
\label{segmentation input}
\end{equation}

Then, $a$ and $b$ can be calculated by $\textbf{s}$ as shown in (\ref{calculate_ab}).

\begin{equation}
a = \textbf{s}M_{1}\textbf{s}^{T}~~~~b = \textbf{s}M_{2}\textbf{s}^{T}
\label{calculate_ab}
\end{equation}
where,
$$ 
M_1 = \left[ \begin{array}{rrrrrrrrr}
1 & 0 & 0 & 0 & 0 & 0 & 0 & 0 & 0\\
0 & \frac{1}{4} & 0 & \frac{1}{2} & 0 & 0 & 0 & 0 & 0\\
0 & 0 & \frac{1}{4} & 0 & 0 & 0 & \frac{1}{2} & 0 & 0\\
0 & 0 & 0 & \frac{1}{4} & 0 & 0 & 0 & 0 & 0\\
0 & 0 & 0 & 0 & 1 & 0 & 0 & 0 & 0\\
0 & 0 & 0 & 0 & 0 & \frac{1}{4} & 0 & \frac{1}{2} & 0\\
0 & 0 & 0 & 0 & 0 & 0 & \frac{1}{4} & 0 & 0\\
0 & 0 & 0 & 0 & 0 & 0 & 0 & \frac{1}{4} & 0\\
0 & 0 & 0 & 0 & 0 & 0 & 0 & 0 & 1\\
\end{array}\right] 
$$
$$ 
M_2 = \left[ \begin{array}{rrrrrrrrr}
0 & 0 & 0 & 0 & 0 & 0 & 0 & 0 & 0\\
0 & \frac{1}{4} & 0 & -\frac{1}{2} & 0 & 0 & 0 & 0 & 0\\
0 & 0 & \frac{1}{4} & 0 & 0 & 0 & -\frac{1}{2} & 0 & 0\\
0 & 0 & 0 & \frac{1}{4} & 0 & 0 & 0 & 0 & 0\\
0 & 0 & 0 & 0 & 0 & 0 & 0 & 0 & 0\\
0 & 0 & 0 & 0 & 0 & \frac{1}{4} & 0 & -\frac{1}{2} & 0\\
0 & 0 & 0 & 0 & 0 & 0 & \frac{1}{4} & 0 & 0\\
0 & 0 & 0 & 0 & 0 & 0 & 0 & \frac{1}{4} & 0\\
0 & 0 & 0 & 0 & 0 & 0 & 0 & 0 & 0\\
\end{array}\right] 
$$

In flow field segmentation methods, scholars usually obtain vortex region by constructing operations related to $a$ and $b$ in (\ref{Frobenius norm}), such as $(b-a)/2$ or $b/(a+b)$ \cite{hunt1987vorticity,liu2016new}. Therefore, by the calculation of vector $\textbf{s}$ in (\ref{segmentation input}), we can get any calculation value of $a$ and $b$, so as to realize the division of the vortex region.

\subsection{Classification criterion}
\label{sc2.2}

In our research, we construct a novel 3D classification criterion by the following arguments, to overcome the problems in section \ref{problem description}.

For flow field $\mathcal{F}$, the momentum equation is expressed as follows \cite{chorin1968numerical}:
\begin{equation}
\begin{aligned}
\nabla \cdot \textbf{v} &= 0 \\
\frac{\partial \textbf{v}}{\partial t} + \textbf{v} \cdot \nabla \textbf{v} &= - \nabla P + \nu~\Delta \textbf{v} \\
\end{aligned}
\label{momentum equation}
\end{equation}
where, $\mathbf{v}$ is the velocity field of fluid, $t$ is time, $P$ is pressure field, $\nabla$ is gradient operator, $\nu$ is the motion viscosity of fluid.

For (\ref{vorticity}), by fixing the coordinate component $z$ in the position $\textbf{p}$, and setting the velocity component $w$ in the velocity $\textbf{v}$ to be 0, the process is expressed by (\ref{slicing process}):
\begin{equation}
\begin{aligned}
\omega_{2D} &= \left. \omega_{3D} \right|_{w = 0,z \in \lbrack z_{0},z_{1},\ldots,z_{n}\rbrack} \\
& = \left. {\left( {\frac{\partial w}{\partial y} - \frac{\partial v}{\partial z}} \right)i + \left( {\frac{\partial u}{\partial z} - \frac{\partial w}{\partial x}} \right)j + \left( {\frac{\partial v}{\partial x} - \frac{\partial u}{\partial y}} \right)k} \right|_{w = 0,z \in \lbrack z_{0},z_{1},\ldots,z_{n}\rbrack} \\
& = \left( {\frac{\partial v}{\partial x} - \frac{\partial u}{\partial y}} \right)k
\end{aligned}
\label{slicing process}
\end{equation}

From (\ref{momentum equation}), we consider the following \ref{2D momentum equation}(a) and \ref{2D momentum equation}(b) are the components of the 2D incompressible momentum equation \cite{chorin1968numerical} in 2D Cartesian coordinate system without external force:
\begin{equation}
\begin{aligned}
\frac{\partial u}{\partial t} + u\frac{\partial u}{\partial x} + v\frac{\partial u}{\partial y} = - \frac{1}{\rho}\frac{\partial P}{\partial x} + \nu\left( {\frac{\partial^{2}u}{\partial x^{2}} + \frac{\partial^{2}u}{\partial y^{2}}} \right)~(a) \\
\frac{\partial v}{\partial t} + u\frac{\partial v}{\partial x} + v\frac{\partial v}{\partial y} = - \frac{1}{\rho}\frac{\partial P}{\partial y} + \nu\left( {\frac{\partial^{2}v}{\partial x^{2}} + \frac{\partial^{2}v}{\partial y^{2}}} \right)~(b) \\
\end{aligned}
\label{2D momentum equation}
\end{equation}

It is common to solve the dimensionless basic equations of fluid mechanics \cite{darwish2016finite}. Define the dimensionless parameters as follows:
$$
\hat{x} = \frac{x}{L}~~~\hat{y} = \frac{y}{L}~~~\hat{u} = \frac{u}{U}~~~\hat{v} = \frac{v}{U}~~~\hat{P} = \frac{P}{\rho U^{2}}~~~\hat{t} = \frac{Ut}{L}
$$
where  $L$ is the characteristic length (usually the diameter of the pipeline), $U$ represents the feature speed, and the $\rho$ is the flow field density.

Thus, (\ref{2D momentum equation}) can be converted into the following dimensionless form:
\begin{equation}
\label{eqw}
\begin{aligned}
& \frac{\partial\hat{u}}{\partial\hat{t}} + \hat{u}\frac{\partial\hat{u}}{\partial\hat{x}} + \hat{v}\frac{\partial\hat{u}}{\partial\hat{y}} = - \frac{\partial\hat{P}}{\partial\hat{x}} + \frac{1}{Re}\left( {\frac{\partial^{2}\hat{u}}{\partial{\hat{x}}^{2}} + \frac{\partial^{2}\hat{u}}{\partial{\hat{y}}^{2}}} \right)~(a) \\
& \frac{\partial\hat{v}}{\partial\hat{t}} + \hat{u}\frac{\partial\hat{v}}{\partial\hat{x}} + \hat{v}\frac{\partial\hat{v}}{\partial\hat{y}} = - \frac{\partial\hat{P}}{\partial\hat{y}} + \frac{1}{Re}\left( {\frac{\partial^{2}\hat{v}}{\partial{\hat{x}}^{2}} + \frac{\partial^{2}\hat{v}}{\partial{\hat{y}}^{2}}} \right)~(b) \\
\end{aligned}
\end{equation}
where $Re = UL/\nu$. Reynolds number (Re) is a dimensionless quantity that helps predict fluid flow patterns in different situations.

Through the following calculation:
$$
\frac{\partial}{\partial\hat{x}} \left[Eq.~5(b)\right] - \frac{\partial}{\partial\hat{y}} \left[Eq.~5(a)\right]
$$

The dimensionless vorticity transport equation can be obtained in (\ref{2D vorticity transport}).

\begin{equation}
\frac{\partial\omega_{2D}}{\partial\hat{t}} = \frac{1}{Re}\left( {\frac{\partial^{2}\omega_{2D}}{\partial{\hat{x}}^{2}} + \frac{\partial^{2}\omega_{2D}}{\partial{\hat{y}}^{2}}} \right) - \left( {\hat{u}\frac{\partial\omega_{2D}}{\partial\hat{x}} + \hat{v}\frac{\partial\omega_{2D}}{\partial\hat{y}}} \right)
\label{2D vorticity transport}
\end{equation}

Then, (\ref{2D vorticity transport}) could be expanded in 3D form as follows:

\begin{equation}
\begin{aligned}
&\left. \frac{\partial\omega_{3D}}{\partial\hat{t}} \right|_{w = 0} = {\bigcup\limits_{z = \lbrack z_{0},z_{1},\ldots,z_{n}\rbrack}\left. \frac{\partial\omega_{3D}}{\partial\hat{t}} \right|_{w = 0}} \\
& = \left. \left( {\frac{1}{Re}\left( {\frac{\partial^{2}\omega_{3D}}{\partial{\hat{x}}^{2}} + \frac{\partial^{2}\omega_{3D}}{\partial{\hat{y}}^{2}}} \right) - \left( {\hat{u}\frac{\partial\omega_{3D}}{\partial\hat{x}} + \hat{v}\frac{\partial\omega_{3D}}{\partial\hat{y}}} \right)} \right) \right|_{w = 0,z = z_{0},z_{1},\ldots,z_{n}}
\end{aligned}
\label{3D vorticity transport}
\end{equation}

The derivation of classification criterion then can be divided into two steps:

\textbf{1)} The relationship between $\partial\omega/\partial t$ and Re.

We start from 2D case. In our research, we fix arbitrary point $\left( {x,y} \right)$ in 2D flow field. Referring to (\ref{2D vorticity transport}), $\hat{u}\left( \partial\omega/\partial\hat{x} \right) + \hat{v}\left( \partial\omega/\partial\hat{y} \right)$ is the convection term which describes the motion of the fluid particle as it moves from one point $\left( {x_0,y_0} \right)$ to another $\left( {x_1,y_1} \right)$. $\partial^{2}\omega/\partial{\hat{x}}^{2}+ \partial^{2}\omega/\partial{\hat{y}}^{2}$ term is the diffusion term. Then Equation (\ref{2D vorticity transport}) represents that $\partial\omega/\partial t$ is controlled by the convection term, the diffusion term and Reynolds number. In our research, we only consider the vorticity change in time dimension, and thus regard the diffusion term and convection term at a fixed point $\left( {x,y} \right)$ as constants. Then we get the following (\ref{classification crterion}) from (\ref{2D vorticity transport}):

\begin{equation}
    \frac{\partial\omega_{2D}}{\partial\hat{t}} \propto \left( \frac{m}{Re} + n \right)
\label{classification crterion}
\end{equation}
where $m$ represents the diffusion term, $n$ represents the convection term, $m$ and $n$ are constants, and $\partial\omega_{2D}/\partial\hat{t}$ represents 2D vorticity change rate with time.

In 3D case, we get the following (\ref{classification criterion_3D}) from (\ref{3D vorticity transport}):

\begin{equation}
\begin{aligned}
&\left. \frac{\partial\omega_{3D}}{\partial\hat{t}} \right|_{w = 0} \propto \left. \left( \frac{m}{Re} + n \right) \right  |_{w = 0,z = z_{0},z_{1},\ldots,z_{n}}
\end{aligned}
\label{classification criterion_3D}
\end{equation}

\textbf{2)} The relationship between Re and vortex wake.

Williamson and Roshko \cite{williamson1988vortex} performed experimental studies of different vortex shedding modes on oscillating cylinders with Reynolds numbers from 300 to 1000. They found that when $Re < 300$, the whole region is dominated by P+S mode, and when $300 < Re < 1000$, the flow field will change from P+S mode to 2P mode at a certain boundary. In Han's work \cite{han2018wake}, as the Reynolds number changes, the main shedding mode of vortex also changes. Previous researches showed that vortex mode could be determined by Reynolds number under specific conditions.

Above analysis shows that $\partial\omega_{2D}/\partial t$ can be employed to differentiate 3D flow fields with different Reynolds numbers and further to classify the types of vortex wakes in 3D flow field.

\section{Methods framework}

\begin{figure*}[!ht]
\centering
\includegraphics[width=\linewidth]{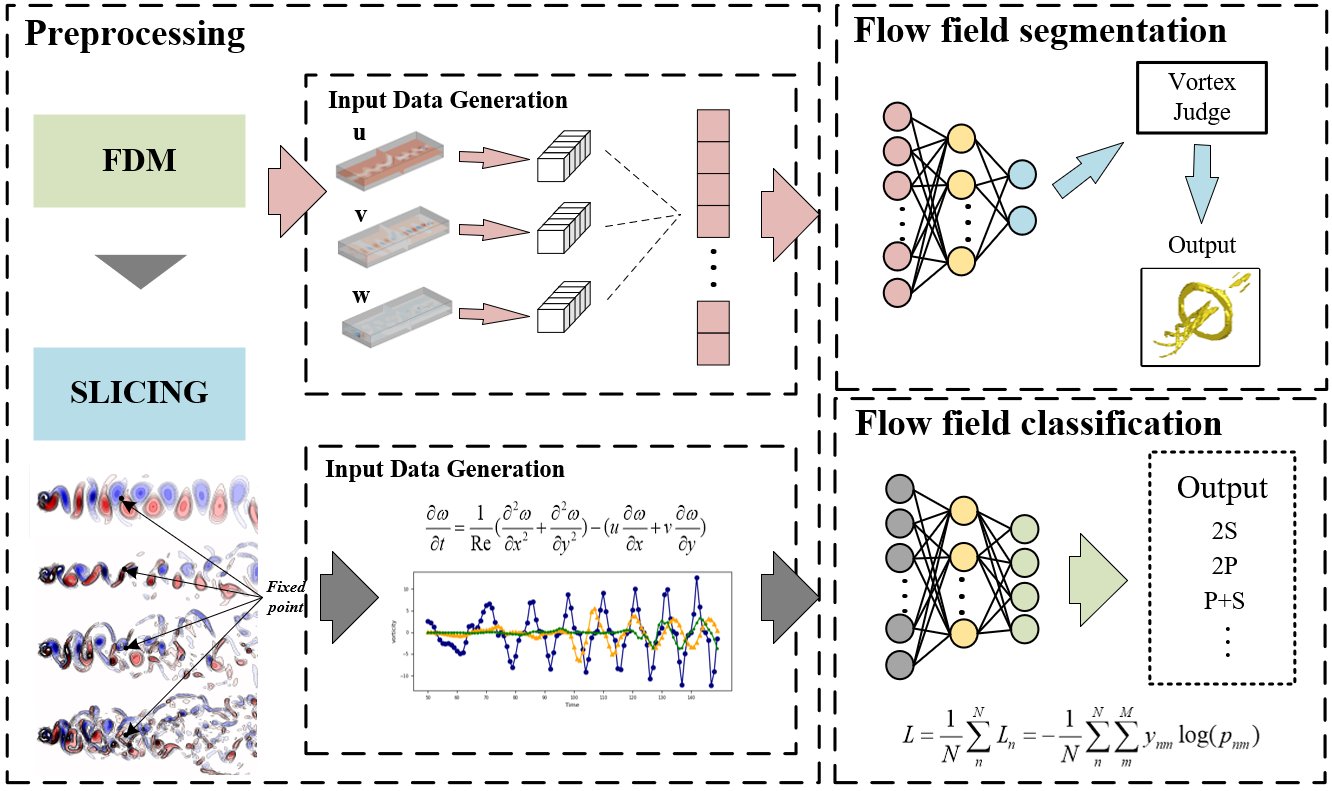}
\caption{The framework of flow field segmentation and classification methods.}
\label{Overall framework}
\end{figure*}

Three problems must be solved in the framework of our methods: 1) generating input data from above segmentation criterion; 2) generating input data according to the mapping between vorticity and vortex wake type from above classification criterion; 3) establishing segmentation and classification networks. The general framework of our methods to solve these three problems are presented below.

The methods framework, as shown in Fig. \ref{Overall framework}, consists of three modules: preprocessing module, flow field segmentation module and flow field classification module. Preprocessing module is to process the continuous data into discrete form and generate data for flow field segmentation and classification modules. Subsequently, flow field segmentation module trains the network to judge the vortex region and perform visualization operations. Finally, flow field classification module classifies flow fields with different Reynolds numbers and map them to corresponding vortex types.

\subsection{Preprocessing module}

The flow field data is provided in form of network common data form (netCDF) \cite{rew1990netcdf}, which contains flow velocity values in each direction of each time step and 3D grid points, so that the data points obtained by sampling are discrete point sets in continuous flow field space. For discrete dataset, we need to redefine the discrete point set and the corresponding velocity and coordinate components of the sampling point. Therefore, define $\mathcal{F}_{d} = \left\{ p_{i,j,k}^{t},i = 0,\ldots,I,j = 0,\ldots,J,k = 0,..,K \right\}$ as the grid point set of 3D flow field $\mathcal{F}$ at time $t$ in 3D Cartesian coordinate system, $t~ = \left\{ ~0,~1,\ldots~,~T \right\}$. Define the components of the coordinate data at the point $p_{i,j,k}^{t}$ in each direction of the 3D Cartesian coordinate system to be $x_{i}, y_{j}, z_{k}$ respectively, and the components of the flow velocity vector in each direction are $u_{i,j,k}^{t}$, $v_{i,j,k}^{t}$, $w_{i,j,k}^{t}$.

The vorticity calculation at discrete coordinate points adopts finite difference method (FDM) \cite{perrone1975general} for spatial discretization, and the central difference form of components in each direction of 3D vorticity at point $p_{i,j,k}^{t}$ is expressed as follows:
\begin{equation}
\begin{aligned}
\frac{\partial w}{\partial y} - \frac{\partial v}{\partial z} &= \frac{w_{i + 1,j,k}^{t} - w_{i - 1,j,k}^{t}}{y_{i + 1} - y_{i - 1}} - \frac{v_{i,j + 1,k}^{t} - v_{i,j + 1,k}^{t}}{z_{j + 1} - z_{j - 1}} \\
\frac{\partial u}{\partial z} - \frac{\partial w}{\partial x} &= \frac{u_{i + 1,j,k}^{t} - u_{i - 1,j,k}^{t}}{z_{i + 1} - z_{i - 1}} - \frac{w_{i,j + 1,k}^{t} - w_{i,j + 1,k}^{t}}{x_{j + 1} - x_{j - 1}} \\
\frac{\partial v}{\partial x} - \frac{\partial u}{\partial y} & = \frac{v_{i + 1,j,k}^{t} - v_{i - 1,j,k}^{t}}{x_{i + 1} - x_{i - 1}} - \frac{u_{i,j + 1,k}^{t} - u_{i,j + 1,k}^{t}}{y_{j + 1} - y_{j - 1}} \\
\end{aligned}
\label{discrete vorticity}
\end{equation}

\subsubsection{Input data generation for segmentation}

As described in section \ref{design_ffs}, the items in vector $\textbf{s}$ in (\ref{segmentation input}) are selected as the input in our method. This allows us to map vector $\textbf{s}$ to two-dimensional vector (vortex and non-vortex), which can be achieved using Multi-layer Perception (MLP).

Through FDM, we fit vector $\textbf{s}$ in each discrete point $x_{i}$, $y_{j}$, $z_{k}$ by selecting data from $u$, $v$ and $w$ according to (\ref{discrete vorticity}). Therefore, our research selects the 15-dimensional vector $\textbf{s}_{i,j,k}$ in $x_{i}$, $y_{j}$, $z_{k}$, as the input as follows:
\begin{equation}
\begin{aligned}
& \textbf{u}_{i,j,k} = \lbrack u_{i,j - 1,k}^{t},~u_{i,j,k}^{t},~u_{i,j + 1,k}^{t},~u_{i,j,k - 1}^{t},~u_{i,j,k + 1}^{t} \rbrack \\
& \textbf{v}_{i,j,k} = \lbrack v_{i - 1,j,k}^{t},~v_{i,j,k}^{t},~v_{i + 1,j,k}^{t},~v_{i,j,k - 1}^{t},~v_{i,j,k + 1}^{t} \rbrack\\
& \textbf{w}_{i,j,k} = \lbrack w_{i - 1,j,k}^{t},~w_{i,j,k}^{t},~w_{i + 1,j,k}^{t},~w_{i,j - 1,k}^{t},~w_{i,j + 1,k}^{t} \rbrack \\
& \textbf{s}_{i,j,k} = \textbf{u}_{i,j,k} \cup \textbf{v}_{i,j,k} \cup \textbf{w}_{i,j,k} \\
\end{aligned}
\label{data generation for seg}
\end{equation}

\subsubsection{Input data generation for classification}
\label{data generation for clas}

Referring to $\omega_{2D}$ described in (\ref{slicing process}) and difference form described in (\ref{discrete vorticity}), the sliced 3D vorticity value $\omega_{2D}$ is obtained.

At a fixed point, $\partial\omega_{2D}/\partial t$ can be learned by learning the periodicity of vorticity through neural network. And by classifying $\partial\omega_{2D}/\partial t$ can realize the classification of 3D flow fields with different Reynolds numbers. 

Therefore, $\omega_{2D}$ at each sampling time $t$ is selected as the input data. According to FDM, the differential form of $\partial\omega_{2D}/\partial t$:
$$\frac{\partial\omega_{2D}}{\partial{t}} = \frac {\omega_{i,j,k}^{t + 1} - \omega_{i,j,k}^{t - 1}}{2\mathrm{\Delta}t}$$

Simultaneously, fixing $j = (J + 1)/2$ (select the points where vorticity changes significantly), our research generates the input vector of neural network in (\ref{input data for clas}).
\begin{equation}
   \textbf{s}_{i,k} = \left\lbrack {\omega_{i,j,k}^{0},\omega_{i,j,k}^{1},\ldots,\omega_{i,j,k}^{T}} \right\rbrack,~j = (J + 1)/2
\label{input data for clas}
\end{equation}

According to the above input vector $\textbf{s}_{i,k}$, deep learning method can learn $\partial\omega_{2D}/\partial t$ at any time $t$.

\subsection{Flow field segmentation module}

The local flow velocity information $\textbf{s}_{i,j,k}$ of each point is obtained through (\ref{data generation for seg}) and input into the corresponding flow field segmentation network respectively. In 3D case, our network is expressed as 
$$
\left\{ \begin{matrix}
{y_{n} = f_{MLP\_ seg}\left( {S_{i,j,k},\theta_{seg}} \right)} \\
{i = 1,\ldots,I,~j = 1,\ldots,J,~k = 1,\ldots,K} \\
{\left. {}\theta_{seg} = \left\{ {W,b} \right\} \right.\sim U\left( {- \sqrt{\frac{3}{l}},\sqrt{\frac{3}{l}}} \right),U\left( {- \sqrt{\frac{1}{l}},\sqrt{\frac{1}{l}}} \right)} \\
\end{matrix} \right.
$$
where $\theta_{seg}$ is the initialization parameter set, $W$ and $b$ are the weight and bias of our network, and $l$ is the amount of input data of each linear layer.

According to network structure, this network uses the $Relu$ function as the activation function. To deal with segmentation problem, the binary cross-entropy loss function is used to train our network, as follows: 
\begin{equation}
Loss\left( {{\hat{y}}_{n},y_{n}} \right) = \frac{1}{N}{\sum\limits_{n}^{N}{- \left\lbrack {y_{n}{\log\left( {\hat{y}}_{n} \right)} + \left( {1 - y_{n}} \right){\log\left( {1 - {\hat{y}}_{n}} \right)}} \right\rbrack}}
\end{equation}
where $N$ represents the number of samples, $n$ represents the sample sequence, $y_n$ represents the predicted value of the $n^{th}$ sample, and ${\hat{y}}_{n}$ represents the true value of the $n^{th}$ sample.

\begin{table*}[!ht]
\centering\small
\begin{tabular}{ccccccc}
\multicolumn{7}{l}{\small{\textbf{Table 1}}}\\
\multicolumn{7}{l}{\small{Summary of training and testing flow field datatsets}}\\
\specialrule{0.05em}{3pt}{3pt}
  &Type &Case &Name  &Grid Size &No. &Exp\\
\specialrule{0.05em}{3pt}{3pt}
\multirow{7}{*}{\makecell{Flow field\\ segmentation}}  &2D &Cylinder  &cylinder2d   &640 × 80 &1 &\uppercase\expandafter{\romannumeral 1}\\
\cmidrule(r){2-7}
    &\multirow{4}{*}{3D} &\multirow{3}{*}{ShockWave}  &SV81-1000  &81 × 81 × 81 &2.1 &\multirow{3}{*}{\uppercase\expandafter{\romannumeral 2}.(1), \uppercase\expandafter{\romannumeral 2}.(2)}\\
    &  & &SV81-1500  &81 × 81 × 81  &2.2\\
    &  & &SV161-2000  &161 × 81 × 81  &2.3\\
\cmidrule(r){3-7}
    &  &Turbulence &turbulence  &65 × 65 × 65  &3 &\uppercase\expandafter{\romannumeral 3}\\
\specialrule{0.03em}{3pt}{3pt}
\multirow{4}{*}{\makecell{Flow field\\ classification}} &\multirow{4}{*}{3D} &\multirow{4}{*}{Half-Cylinder} &half-cylinder160  &640 × 240 × 80  &4.1 &\multirow{4}{*}{\uppercase\expandafter{\romannumeral 4}.(1), \uppercase\expandafter{\romannumeral 4}.(2)}\\
& & &half-cylinder320  &640 × 240 × 80  &4.2\\
& & &half-cylinder640  &640 × 240 × 80  &4.3\\
& & &half-cylinder6400  &640 × 240 × 80  &4.4\\
\specialrule{0.05em}{2pt}{0pt}
\end{tabular}
\end{table*}

\subsection{Flow field classification module}

After preprocessing the local calculation of vorticity in (\ref{input data for clas}), vector $\textbf{s}_{i,k}$ is input into multi-layer perceptron for learning, 
$$
\left\{ \begin{matrix}
{y_{n} = f_{MLP\_clas}\left( {\textbf{s}_{i,k},\theta_{clas}} \right)} \\
{i = 1,\ldots,I,~k = 1,\ldots,K} \\
{\left. {}\theta_{clas} = \left\{ {W,b} \right\} \right.\sim U\left( {- \sqrt{\frac{3}{l}},\sqrt{\frac{3}{l}}} \right),U\left( {- \sqrt{\frac{1}{l}},\sqrt{\frac{1}{l}}} \right)} \\
\end{matrix} \right.
$$
where $y_n$ is the output value obtained from the $n^{th}$ selected point in 3D flow field, $\theta_{clas}$ is the initialization parameter set, $W$ and $b$ are the weight and bias of this model, and $l$ is the amount of input data of each linear layer.

The multi-classification cross-entropy loss function is used to train our model. Besides, the Adam optimizer is used to iteratively update the network parameters. The form of our loss function is expressed as follow:
\begin{equation}
\begin{aligned}
Loss\left( {{\hat{y}}_{n},y_{n}} \right) = \frac{1}{N}{\sum\limits_{n}^{N}{L_{n}\left( {\hat{y}}_{n},y_{n} \right)}} = - \frac{1}{N}{\sum\limits_{n}^{N}{\sum\limits_{m}^{M}{y_{nm}{\log\left( p_{nm} \right)}}}}
\end{aligned}
\end{equation}
where $N$ represents the number of samples, $M$ represents the number of categories, $n$ represents the sample sequence, $m$ represents the vector number, $y_n$ represents the predicted value of the $n^{th}$ sample, and ${\hat{y}}_{n}$ represents the true value of the $n^{th}$ sample. $y_{nm}$ is a sign function (0 or 1), and if the category $m$ of the sample $n$ is the real category, the sign function takes 1, otherwise it takes 0. $p_{nm}$ is the predicted probability that the observed sample belongs to category $m$.

\section{Results and discussions}

\subsection{Data and Experimental Setup}

This section introduces the datasets required for experiment and its related parameters. Then, prepare several sets of experiments separately for flow field segmentation and classification.

\subsubsection{Datasets}

To verify the performance of our flow field segmentation and classification method, we utilize several 2D and 3D cases as task datasets.

In segmentation task, cases are selecteted as follows:

\textbf{\emph{Cylinder}} \cite{gunther2017generic}: This public dataset is derived from the simulation of a viscous 2D flow around a cylinder using the Gerris flow solver \cite{popinet2004free}. The regular grid resolution is 640 × 80 and the Reynolds number is 160.

\textbf{\emph{Shockwave (SV)}}: This 3D datasets consists of three data sets: SV81-1000, SV81-1500 and SV161-2000. The grid resolution of SV81-1000 and SV81-1500 are 81 × 81 × 81 with their Reynolds numbers are 1000 and 1500 respectively. The grid resolution of SV161-2000 is 161 × 81 × 81 with Reynolds numbers is 2000.

\textbf{\emph{Turbulence}}: This 3D dataset is derived from the simulation of a viscous 3D flow and the regular grid resolution is 65 × 65 × 65.

\begin{figure}[!ht]
\centering
\includegraphics[width=\linewidth]{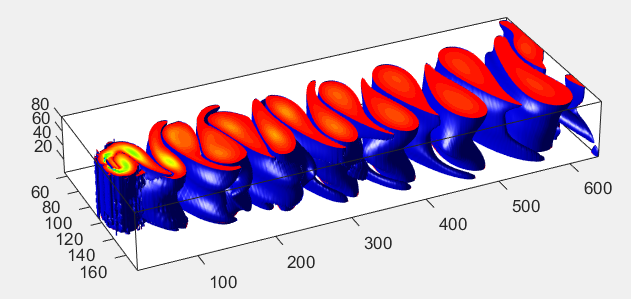}
\caption{3D isosurface map of Half Cylinder dataset when Re = 160}
\label{isosurface}
\end{figure}

In classification task, cases are selecteted as follows:

\textbf{\emph{Half Cylinder}} \cite{rojo2019vector}: This public dataset is a small collection of simulated incompressible 3D flows around a half cylinder generated using the Gerris flow solver \cite{popinet2004free}, where the Reynolds number ranges from 160 to 6400 (Re: 160, 320, 640, and 6400). The regular grid resolution is 640 × 240 × 80, and the data is provided in the network common data table (netCDF) format \cite{rew1990netcdf}. The 3D isosurface diagram of this flow field at a Reynolds number of 160 is shown in Fig. \ref{isosurface}.

In Table 1, we summarize the dataset information of the flow field segmentation and flow field classification experiments, and their related parameters.

\subsubsection{Experimental Setup}

As shown in Table 1, we set up the following sets of experiments based on the above datasets.

In segmentation task:

\textbf{\emph{Cylinder}}: In Exp \uppercase\expandafter{\romannumeral 1}, this experiment randomly divide sampling points in dataset 1 by 8:2 and obtain 40,960 training data and 10,240 test data.

\textbf{\emph{Shockwave (SV)}}: In the case of shockwave, two different experiments are performed to validate our method. In Exp \uppercase\expandafter{\romannumeral 2}.(1), dataset 2.1 with a size of 81 × 81 × 81 is used as the training set and Dataset 2.2 with a size of 81 × 81 × 81 is used as the testing set. In Exp \uppercase\expandafter{\romannumeral 2}.(2), after randomly dividing dataset 2.1 and dataset 2.3 by 8:2, these two dataset are combined in a ratio of 8:2 to obtain training set and test set.

\textbf{\emph{Turbulence}}: In Exp \uppercase\expandafter{\romannumeral 3}, this experiment randomly divide sampling points in dataset 3 by 8:2 and obtain 219,700 training data and 54,925 test data.

\begin{figure*}
\centering
\subfigure[Vortex zone labeled by IVD method]{
\includegraphics[width=0.7\linewidth]{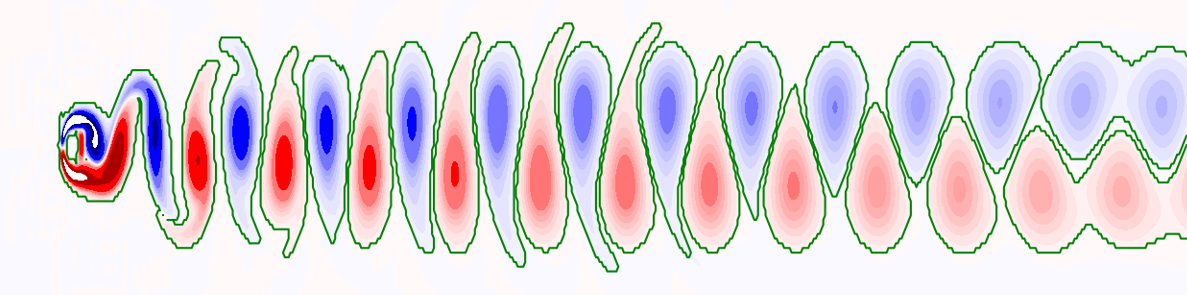}
\label{Fig3-1}
}
\quad
\subfigure[Flow field segmentation result]{
\includegraphics[width=0.7\linewidth]{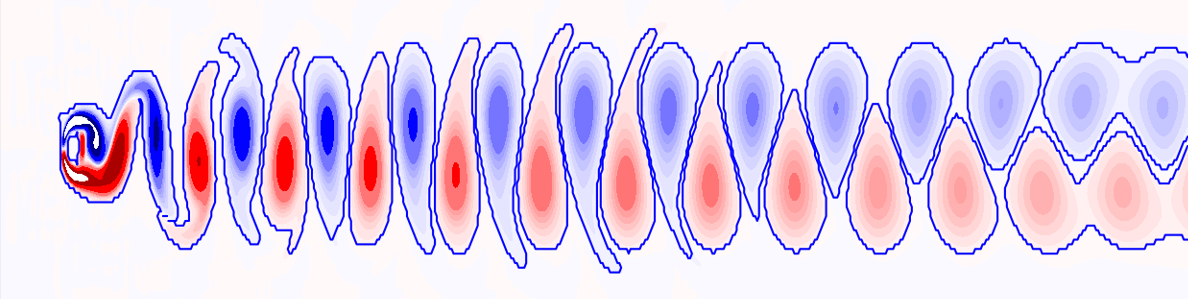}
\label{Fig3-2}
}
\quad
\caption{The visual results in Exp \uppercase\expandafter{\romannumeral 1} on Cylinder case.}
\label{cylinder2d}
\end{figure*}

In flow field classification experiment, as shown in Table 1, we set up the following sets of experiments.

\textbf{\emph{Half Cylinder}}: In Exp \uppercase\expandafter{\romannumeral 4}.(1) and Exp \uppercase\expandafter{\romannumeral 4}.(2), we select datasets 4.1, 4.2 and datasets 4.1, 4.2, 4.3, 4.4 respectively. 

Through dimensionality reduction, 80 2D flow field cutting planes can be selected from each 3D dataset. In each cut plane, 640 input vectors can be obtained through data selection. For cross-validation, we divide each 3D dataset into 5 groups with each group has 16 2D cut plane. For each group, 80\% of the data is selected for training, and the remaining 20\% is selected for testing. That is, 5 groups of experiments can be carried out in Exp \uppercase\expandafter{\romannumeral 4}.(1) and Exp \uppercase\expandafter{\romannumeral 4}.(2).

\subsection{Benchmark methods and Metrics}

In flow field segmentation experiments, our method is compared with four other techniques, which are defined as follows:

\textbf{\emph{Vortex-Net}} \cite{deng2019cnn}: a segmentation CNN to detect vortices from the velocity field.

\textbf{\emph{Vortex-Seg-Net}} \cite{wang2021rapid}: a fully convolutional method for flow field segmentation.

\textbf{\emph{U-Net}}: 3D eddy detection method based on U-Net.

\textbf{\emph{IVD}} \cite{haller2016defining}: a threshold-independent vortex detection technique based on the outermost closed-level sets of the IVD value.

In flow field classification experiments, our method is compared with three common image processing methods, which are defined as follows:

\textbf{\emph{CNN}} \cite{ribeiro2022machine}: a common convolutional method suitable for image classification.

\textbf{\emph{FCN}}: a fully convolutional method modified for classification task.

\textbf{\emph{U-Net}}: a classification network based on U-Net modified for classification task.

The performances for segmentation and classification methods are measured by precision, recall, accuracy and time consumption. Precision and recall rate are as shown below:
$$
Precision = \frac{TP}{TP + FP}
$$
$$
Recall = \frac{TP}{TP + FN}
$$
where $TP$, $FP$, $TN$, and $FN$ mean the number of true positives, false positives, true negatives, and false negatives. 

\subsection{Experimental results}

\subsubsection{Results for flow field segmentation}

\begin{figure*}[!ht]
\centering
\subfigure[Vortex-Net method]{
\includegraphics[width=0.3\linewidth]{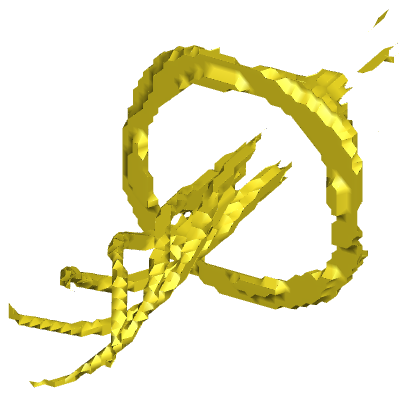}
\label{Fig4-1}
}
\quad
\subfigure[Vortex-Seg-Net method]{
\includegraphics[width=0.3\linewidth]{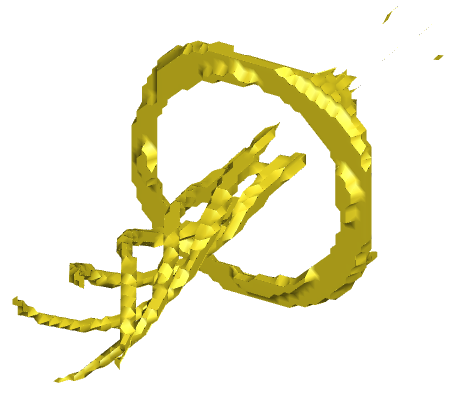}
\label{Fig4-2}
}
\subfigure[U-Net method]{
\includegraphics[width=0.3\linewidth]{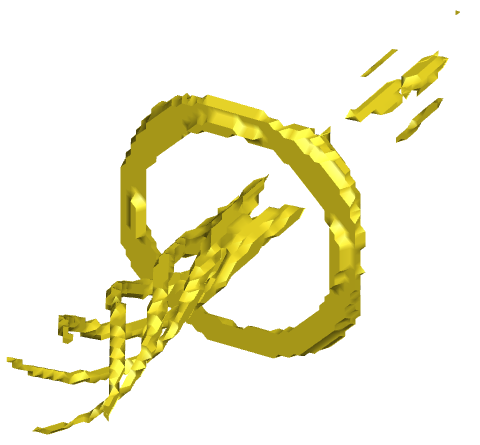}
\label{Fig4-3}
}
\quad
\subfigure[MLP method]{
\includegraphics[width=0.3\linewidth]{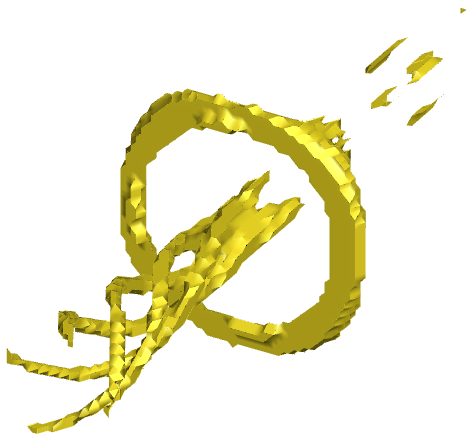}
\label{Fig4-4}
}
\quad
\subfigure[IVD method]{
\includegraphics[width=0.3\linewidth]{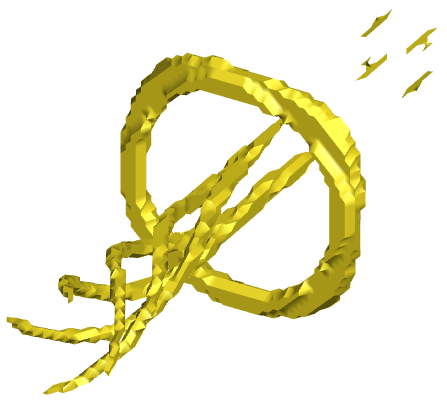}
\label{Fig4-5}
}
\quad
\caption{The visual results of different methods in Exp \uppercase\expandafter{\romannumeral 2}.(1) on ShockWave case.}
\label{SV3D}
\end{figure*}

The hyper-parameters for flow field segmentation experiment are summarized in Table 2. 

\begin{table}[!tb]
\centering\small
\begin{tabular}{lllll}
\multicolumn{5}{l}{\small{\textbf{Table 2}}}\\
\multicolumn{5}{l}{\small{Hyperparameters for the flow field segmentation experiments}}\\
\specialrule{0.05em}{3pt}{3pt}
Type &\makecell{MLP \\ Exp \uppercase\expandafter{\romannumeral 1}} &\makecell{MLP \\ Exp \uppercase\expandafter{\romannumeral 2}.(1)} &\makecell{MLP \\ Exp \uppercase\expandafter{\romannumeral 2}.(2)} &\makecell{MLP \\ Exp \uppercase\expandafter{\romannumeral 3}}\\
\specialrule{0.05em}{3pt}{3pt}
Input           &6    &15  &15  &15\\
Hidden layer1   &128  &64  &64  &64\\
Hidden layer2   &128  &64  &64  &64\\
Output          &2    &2   &2   &2\\
\specialrule{0.05em}{3pt}{3pt}
Learning rate   &0.005 &0.005 &0.005 &0.005\\
Epoch           &500   &500   &500   &500\\
Batch size(Training)   &128  &19,683  &39,123  &4,225\\
Batch size(Testing)  &1,024   &19,683  &19,683  &4,225\\
\specialrule{0.05em}{2pt}{0pt}
\end{tabular}
\end{table}

Each indicator is the average calculation result obtained after repeated training 10 times under different random seeds. Table 4 presents the accuracy and time-consuming performance of all mainstreaming methods in flow field segmentation experiments.

\begin{table}[!ht]
\centering\small
\begin{tabular}{llllc}
\multicolumn{5}{l}{\small{\textbf{Table 3}}}\\
\multicolumn{5}{l}{\small{Experiment results for flow field segmentation}}\\
\specialrule{0.05em}{3pt}{3pt}
Experiment &Method &Precision  &Recall &\makecell{Time \\ Consumption \\ (s)}\\
\specialrule{0.05em}{3pt}{3pt}
\multirow{2}{*}{Exp \uppercase\expandafter{\romannumeral 1}} &MLP   &\textbf{99.30}\%  &\textbf{98.87}\%  &\textbf{2.3}\\
    &IVD   &100\%  &100\% &432\\
\specialrule{0.03em}{3pt}{3pt}
\multirow{5}{*}{Exp \uppercase\expandafter{\romannumeral 2}.(1)}  &Vortex-Net &76.68\% &\textbf{89.25}\% &565\\
    &Vortex-Seg-Net  &82.26\%  &81.16\%  &43.2\\
    &U-Net  &86.17\%  &84.25\%  &22.3\\
    &MLP   &\textbf{90.76}\%  &82.73\%  &\textbf{4.5}\\
    &IVD   &100\%  &100\% &3023\\
\specialrule{0.03em}{3pt}{3pt}
\multirow{2}{*}{Exp \uppercase\expandafter{\romannumeral 2}.(2)}  
    &MLP   &\textbf{90.77}\%  &\textbf{84.30}\%  &\textbf{6.1}\\
    &IVD   &100\%  &100\% &4435\\
\specialrule{0.03em}{3pt}{3pt}
\multirow{5}{*}{Exp \uppercase\expandafter{\romannumeral 3}} &Vortex-Net &85.22\%  &95.23\%  &76.2\\
&Vortex-Seg-Net &92.23\%  &97.06\%  &20.1\\
&U-Net &97.86\%  &\textbf{98.14}\%  &12.1\\
&MLP &\textbf{98.87}\%  &97.92\%  &\textbf{4.7}\\
&IVD &100\%  &100\%  &3879\\
\specialrule{0.05em}{2pt}{0pt}
\end{tabular}
\end{table}

\textbf{\emph{Cylinder}}: In Exp \uppercase\expandafter{\romannumeral 1}, compared with the label area, its accuracy rate is as high as 99.40\%. At the same time, the precision and recall rate under this experiment respectively reaches 99.30\% and 98.87\%.

\begin{figure}[!ht]
\label{clas_result2class}
\centering
\quad
\subfigure[test accuracy]{
\includegraphics[width=\linewidth]{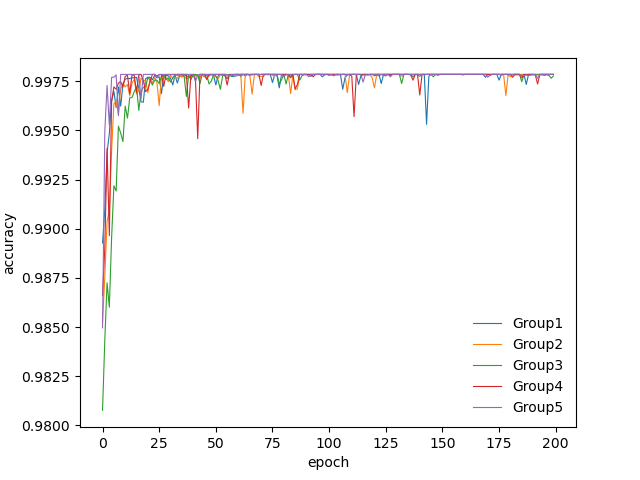}
\label{classification_accuracy_2class}
}
\quad
\subfigure[test loss]{
\includegraphics[width=\linewidth]{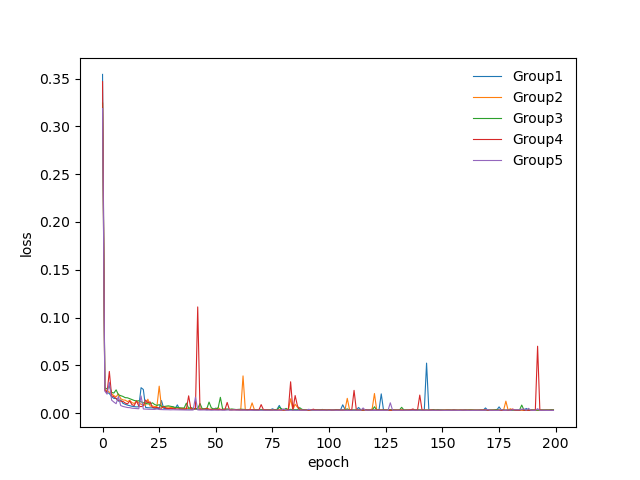}
\label{classification_loss_2class}
}
\centering
\caption{The training loss and test loss and accuracy curve of our method in Exp \uppercase\expandafter{\romannumeral 4}.(1).}
\end{figure}

\begin{figure}[!ht]
\label{clas_result4class}
\centering
\quad
\subfigure[test accuracy]{
\includegraphics[width=\linewidth]{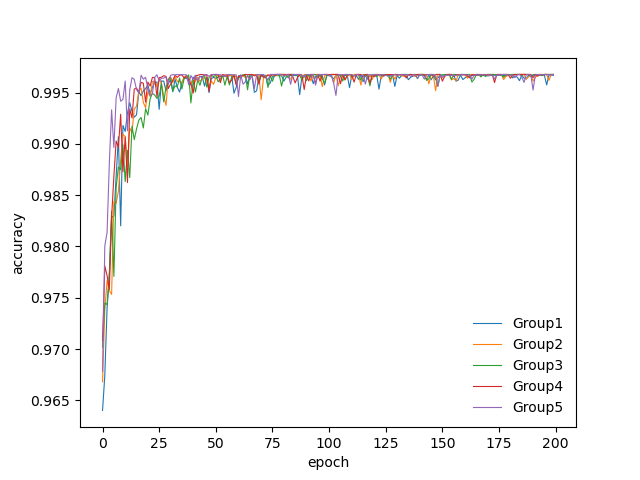}
\label{classification_accuracy_4class}
}
\quad
\subfigure[test loss]{
\includegraphics[width=\linewidth]{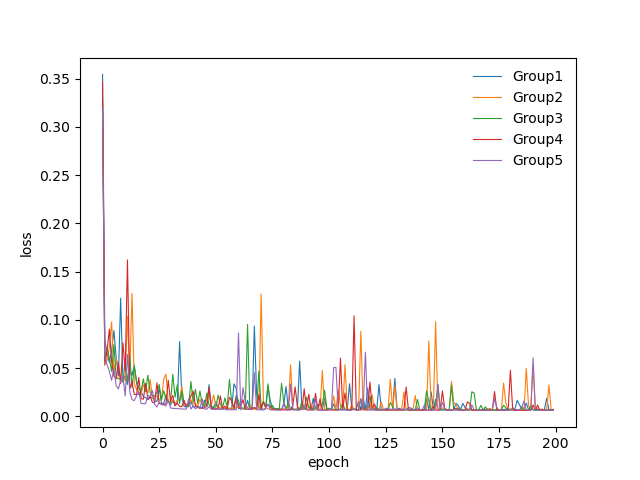}
\label{classification_loss_4class}
}
\centering
\caption{The training loss and test loss and accuracy curve of our method in Exp \uppercase\expandafter{\romannumeral 4}.(2).}
\end{figure}

\begin{figure*}[!ht]
\centering
\quad
\subfigure[Group 1]{
\includegraphics[width=7cm]{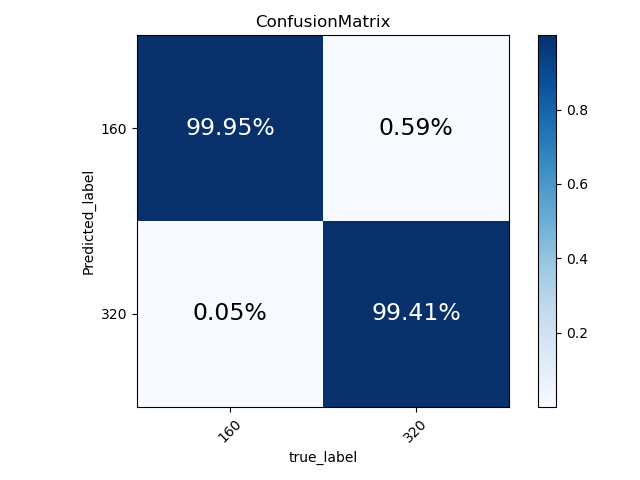}
\label{Fig7-1}
}
\quad
\subfigure[Group 2]{
\includegraphics[width=7cm]{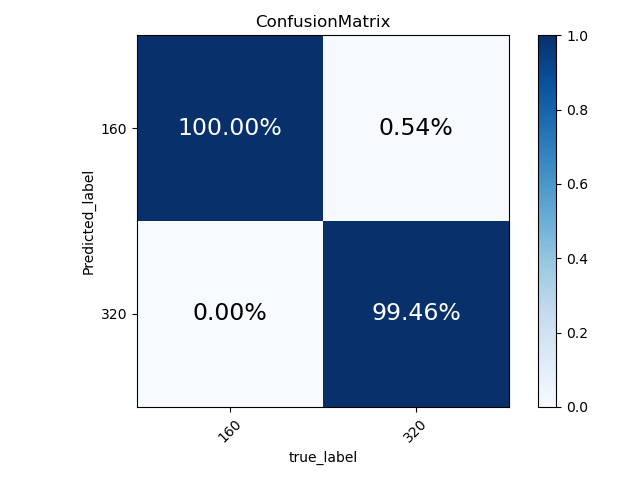}
\label{Fig7-2}
}
\centering
\caption{Confusion matrixes in Exp \uppercase\expandafter{\romannumeral 4}.(1).}
\label{confusion_2class}
\end{figure*}

\begin{figure*}[!ht]
\centering
\quad
\subfigure[Group 1]{
\includegraphics[width=7cm]{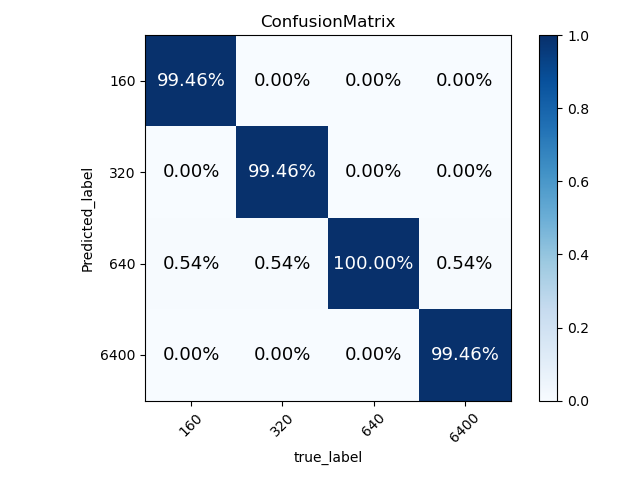}
\label{Fig8-1}
}
\quad
\subfigure[Group 2]{
\includegraphics[width=7cm]{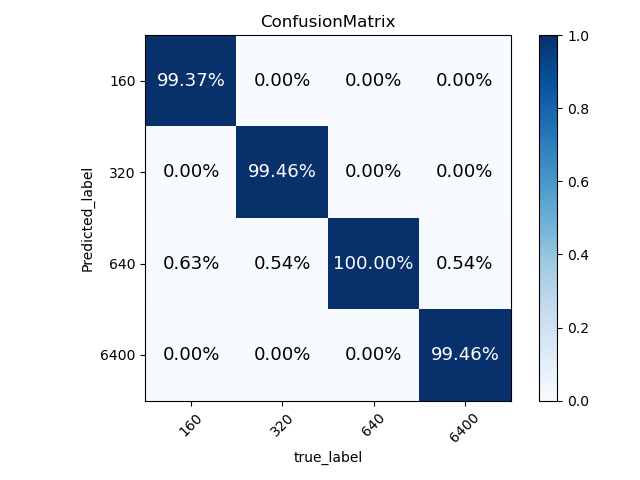}
\label{Fig8-2}
}
\centering
\caption{Confusion matrixes in Exp \uppercase\expandafter{\romannumeral 4}.(2).}
\label{confusion_4class}
\end{figure*}

Experimental results verify the applicability of our method for flow field segmentation in the case of 2D cylinder.

\textbf{\emph{Shockwave}}: In Exp \uppercase\expandafter{\romannumeral 2}.(1) and Exp \uppercase\expandafter{\romannumeral 2}.(2), the experiments respectively achieve 99.75\% and 99.77\% accuracy. From Table 3, compared with the best result (U-Net) in the baseline methods, the precision of our method is 4\% higher in Exp \uppercase\expandafter{\romannumeral 2}.(1). In terms of recall rate, the results of Exp \uppercase\expandafter{\romannumeral 2}.(1) show that our method has a small decrease of 1.5\% and 6\% compared with U-Net method and CNN method. In terms of time consumption, the time of our method is 18 seconds less than that of the U-Net method, and it is also much shorter than label method. In Exp \uppercase\expandafter{\romannumeral 2}.(2), other methods cannot be used as comparative experiments due to the experimental setup. Our method achieved a precision of 90.77\% and a recall rate of 84.30\% in Exp \uppercase\expandafter{\romannumeral 2}.(2). What's more, the consumption time of our method is much lower than IVD method.

\textbf{\emph{Turbulence}}: In Exp \uppercase\expandafter{\romannumeral 3}, the turbulence case is tested by multiple methods. Compared with the U-Net method, our method is 1\% higher in accuracy. In terms of time consumption, this method is improved by 8 seconds compared with the U-Net method.

In the above comparison experiments, our method greatly reduces time consumption in Exp \uppercase\expandafter{\romannumeral 2}.(1) and Exp \uppercase\expandafter{\romannumeral 3} compared with the best benchmark method (U-Net). To shockwave case and turbulence case, the precision of our method is higher than that of U-Net method while ensuring low time consumption. In conclusion, compared with traditional numerical calculation methods, such as IVD, our method can complete the segmentation task in a short time while ensuring high accuracy. Compared with deep learning-based methods such as U-Net, our method has a simpler structure and more efficient segmentation capabilities, so the verification of our method takes less time. Simultaneously, our method is a single-point judgment method based on local information, and its judgment accuracy in ShockWave case and turbulence case is also higher than that of methods based on global information.

Then use tecplot to draw the detection results to observe the vortex region more intuitively.

In Exp \uppercase\expandafter{\romannumeral 1}, the vortex division of 2D cylinder flow field is shown in Fig. \ref{cylinder2d}. In Fig. \ref{cylinder2d}(a), the global detection method IVD \cite{haller2016defining} demarcates the labeled vortex region. Fig. \ref{cylinder2d}(b) shows the vortex area divided by our method.  In addition, the center area of the vortex around the cylinder is perfectly divided, and the wrong part is only the edge of the vortex.

In Exp \uppercase\expandafter{\romannumeral 2}.(1), the visualization results of all methods in SV case are shown in Fig. \ref{SV3D}. Although in Fig. \ref{SV3D}(d), there are some false detections and missing detections in the visulization result of our method. Compared with several other methods in Fig. \ref{SV3D}(a), \ref{SV3D}(b) and \ref{SV3D}(c), our method obtains the closest visualization result. Our method divides the general structure of shockwave, and the segmentation accuracy reaches 99.76\%. The recognition ability of our method in 3D shockwave case is verified.

The above conclusions and visualization results prove that the MLP-based segmentation method proposed in this paper is effective for the flow field segmentation task. Our method is a low-time-consuming and high-accuracy method.

\subsubsection{Results for flow field classification}

To reduce the impact of random division, each indicator is the average calculation result obtained after repeated training 10 times under different random seeds.

The change curves of accuracy rate and loss value of our method in Exp \uppercase\expandafter{\romannumeral 4}.(1) and Exp \uppercase\expandafter{\romannumeral 4}.(2) are respectively shown in Fig. \ref{classification_accuracy_2class} and Fig. \ref{classification_loss_2class}, Fig. \ref{classification_accuracy_4class} and Fig. \ref{classification_loss_4class}. In these figures, five sets of experiments fleetly converge to the same value in terms of accuracy and loss, which proves the reproducibility of our method on half-cylinder case.

The confusion matrices of randomly 2 times in 5 times experiments are shown in Fig. \ref{confusion_2class} and Fig. \ref{confusion_4class}. For classification failure cases, we track the proportion of cases assigned to each error Reynolds number type. The confusion matrix describes the results for all combinations. The diagonal entries of these matrices show the accuracy with which the network correctly classified the dataset. In the cases of misclassification, off-diagonal entries quantify the degree of confusion between classes.

The confusion matrices in Fig. \ref{confusion_2class} show that, the prediction accuracy are 99.95\% and 100\% for the Reynolds number 160 dataset and 99.41\% and 99.46\% for the Reynolds number 320 dataset in Exp \uppercase\expandafter{\romannumeral 4}.(1). The confusion matrices in Fig. \ref{confusion_4class} show that in Exp \uppercase\expandafter{\romannumeral 4}.(2), the prediction accuracy are 99.46\% and 99.37\% for the Reynolds number 160 dataset, 99.46\% and 99.46\% for the Reynolds number 320 dataset, 100\% and 100\% for the Reynolds number 640 dataset , and the prediction accuracy for the dataset with Reynolds number 6400 are both 99.46\%. Clearly, in all combinations, the accuracy of predicting the correct class is above 99\%, up to 100\%.

\begin{table}[!ht]
\centering\small
\resizebox{1.0\linewidth}{!}{
\begin{tabular}{lccc}
\multicolumn{4}{l}{\small{\textbf{Table 4}}}\\
\multicolumn{4}{l}{\small{Experiment result for flow field classification}}\\
\specialrule{0.05em}{3pt}{3pt}
Experiment &Method &Accuracy &Time Consumption(s)\\
\specialrule{0.05em}{3pt}{3pt}
\multirow{4}{*}{Exp \uppercase\expandafter{\romannumeral 4}.(1)} &CNN   &\textbf{99.84}\%   &39.91\\
    &FCN    &99.77\%  &39.72\\
    &U-Net  &99.79\% &49.76\\
    &MLP    &99.78\%  &\textbf{0.40}\\
\specialrule{0.03em}{3pt}{3pt}
\multirow{4}{*}{Exp \uppercase\expandafter{\romannumeral 4}.(2)} &CNN   &99.72\%   &66.45\\
    &FCN    &99.71\%  &67.54\\
    &U-Net  &\textbf{99.76}\% &74.64\\
    &MLP    &99.64\%  &\textbf{0.79}\\
\specialrule{0.05em}{2pt}{0pt}
\end{tabular}}
\end{table}

Fig. \ref{Classification result} shows the intuitive comparison of all methods in Exp \uppercase\expandafter{\romannumeral 4}.(1) and Exp \uppercase\expandafter{\romannumeral 4}.(2) in terms of accuracy rate and training time. The following results can be obtained from Table 4:

\begin{figure*}[!ht]
\centering
\quad
\subfigure[accuracy]{
\includegraphics[width=8.5cm]{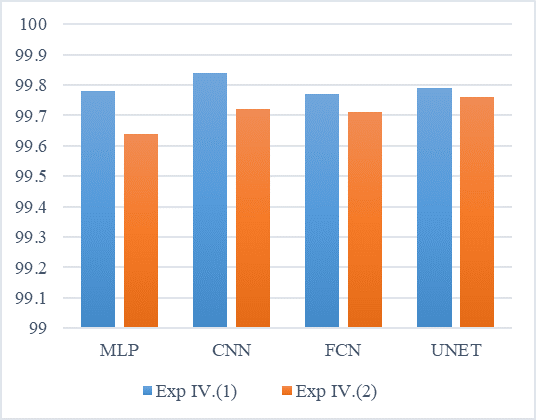}
\label{Fig9-1}
}
\quad
\subfigure[time consumption]{
\includegraphics[width=8.5cm]{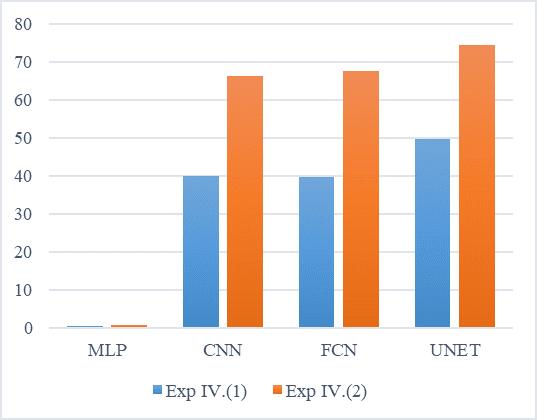}
\label{Fig9-2}
}
\centering
\caption{A straight-square comparison plot of experimental results of each classification method in Exp \uppercase\expandafter{\romannumeral 4}.}
\label{Classification result}
\end{figure*}

1) In Exp \uppercase\expandafter{\romannumeral 4}.(1), CNN, FCN and U-Net respectively achieve 99.84\%, 99.77\%, and 99.79\% accuracy. Our method achieves 99.78\% in this experiment, which is 0.06\% lower than the best performance (CNN) among baseline methods. In Exp \uppercase\expandafter{\romannumeral 4}.(2), CNN, FCN and U-Net respectively achieve 99.72\%, 99.71\%, and 99.76\% accuracy. Our method achieves 99.64\%, which is 0.12\% lower than the best performance (U-Net) among baseline methods.

2) In terms of time overhead, CNN, FCN and U-Net respectively cost 39.91 and 66.45 seconds, 39.72 and 67.54 seconds, 49.76 and 74.64 seconds in Exp \uppercase\expandafter{\romannumeral 4}.(1) and Exp \uppercase\expandafter{\romannumeral 4}.(2), which is quite time-consuming. Our method achieves a time consumption of 0.40 and 0.79 seconds, which is 98\% less than the best method (FCN, CNN) among baseline methods.

From the above results, we draw the following conclusions:

1) In terms of accuracy, our classification method based on local vorticity changes is slightly inferior to the classification performance of the image-based method. Affected by the sample size of our method, there are 10,240 vorticity change vectors collected in each class in Exp \uppercase\expandafter{\romannumeral 4}.(1) and Exp \uppercase\expandafter{\romannumeral 4}.(2), but the number of image samples collected by the unified dataset is only 1,800 in each class. In this case, our method can still achieve a classification effect of more than 99\%, which proves the rationality of our method.

2) In terms of time consumption, our method discards complex networks in favor of simple linear network for classification. Compared with the popular high-complexity methods (CNN, FCN, UNet), it can greatly simplify the training process and reduce training and verification time.

The above conclusions prove that our classification method proposed in this paper is effective for 3D flow field classification task.

\section{Conclusions}

The conclusions of our work are as follows:

1) form a robust 3D flow field segmentation method by mapping local velocity information of sample point to the vortex area through deep learning network;

2) construct a novel 3D flow field classification method by establishing the relationship between vorticity, Reynolds number and vortex wake type through deep learning network;

3) compared with Vortex-Net \cite{deng2019cnn}, Vortex-Seg-Net \cite{wang2021rapid} and U-Net, the precision of vortex region segmented by our method is significantly improved in some cases while consuming over 50\% less time;

4) compared with CNN, FCN and U-Net, the time consumption for flow field classification consumed by our method is reduced by over 98\% while keeping the same predicted accuracy.

Our method can be further improved in some aspects. First, we did not conduct research on flow field segmentation in unstructured grids, but only focused on the segmentation of flow fields in structured grids, which limited the application of our method. Besides, our classification method is not extended to turbulent flow, which lacked certain generalization ability. Since the flow field classification method has advantages in the classification of local Reynolds numbers, it can be applied to turbulence identification in the future.

\section*{Acknowledgement}

This work is supported by the National Natural Science Foundation of China (NSFC) [Grant no. 62250067]. The authors also wish to thank Dr. Liwei Hu from University of Electronic Science and Technology of China (UESTC), for discussions on mathematical modelling.

\end{document}